\documentclass[10pt,twocolumn,letterpaper]{article}

\usepackage[final]{iccv}      % To produce the REVIEW version
\definecolor{iccvblue}{rgb}{0.21,0.49,0.74}
\usepackage[pagebackref,breaklinks,colorlinks,allcolors=iccvblue]{hyperref}

\def\paperID{2123} % *** Enter the Paper ID here
\def\confName{ICCV}
\def\confYear{2025}

\usepackage{float}

\usepackage[utf8]{inputenc} % allow utf-8 input
\usepackage[T1]{fontenc}    % use 8-bit T1 fonts
\usepackage{url}            % simple URL typesetting
\usepackage{booktabs}       % professional-quality tables
\usepackage{amsfonts}       % blackboard math symbols
\usepackage{nicefrac}       % compact symbols for 1/2, etc.
\usepackage{microtype}      % microtypography
\usepackage{graphicx}
\usepackage{dblfloatfix}
\usepackage{graphicx}
\usepackage{caption}
\usepackage{multirow}
\usepackage[accsupp]{axessibility}  % Improves PDF readability for those with disabilities.

\definecolor{citecolor}{HTML}{0071BC}
\definecolor{linkcolor}{HTML}{ED1C24}

\newcommand{\minisection}[1]{\vspace{0.005in} \noindent {\bf #1}}

\newcommand{\quotes}[1]{``#1''}

\def\ourmethod{{\textit{ISLock}}\xspace}
\def\ouratm{{\textit{ATM}}\xspace}
\def\ouracr{{\textit{AdaCR}}\xspace}

\title{\textit{Anchor Token Matching:} \\ Implicit Structure Locking for  Training-free AR Image Editing \vspace{-5mm}}
\author{
    Taihang Hu$^{1}$\footnotemark[1], 
    Linxuan Li$^{1}$\footnotemark[1],
    Kai Wang$^{3,4}$\footnotemark[2] ,
    Yaxing Wang$^{2,1}$\footnotemark[2] ,
    Jian Yang$^{1}$,
    Ming-Ming Cheng$^{2,1}$ \\
    $^1${VCIP, College of Computer Science, Nankai University}, $^2${NKIARI, Shenzhen Futian}\\
    $^3${Computer Vision Center, Universitat Autònoma de Barcelona} \\
    $^4${City University of Hong Kong (Dongguan)} \\
    \texttt{\{hutaihang00, linxuanli520\}@gmail.com}\\
    \texttt{kai.wang@cityu-dg.edu.cn},
    \texttt{\{csjyang,cmm,yaxing\}@nankai.edu.cn}
}

\begin{document}
% \maketitle
\twocolumn[{%v
\maketitle
% \vspace{-1em}
% \vspace{-1em}
\begin{center}
    \centering
    \vspace{-7mm}
    \includegraphics[width=0.999\linewidth]{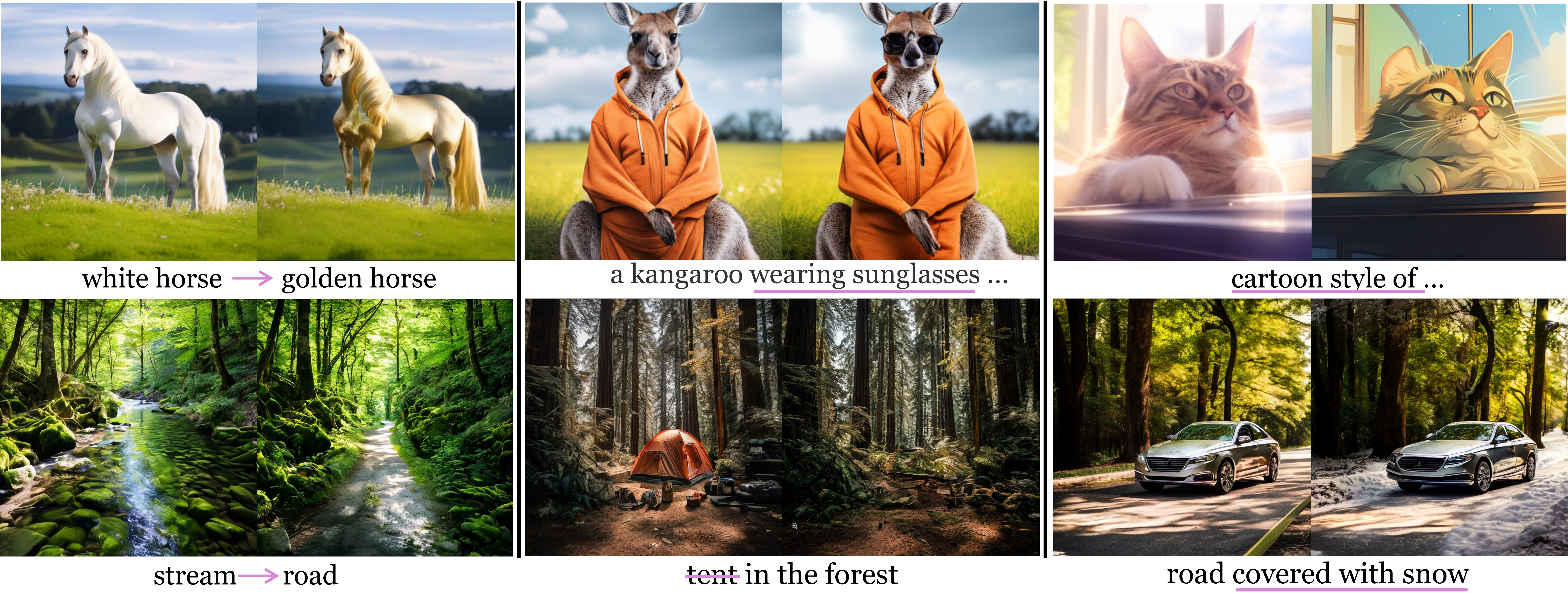}
    \vspace{-7mm}
    \captionof{figure}{
    Our method, Implicit Structure Locking (\ourmethod), is able to achieve the attribute/object replacement (left), add/remove object (mid) and style/state transfer (right) tasks while maintaining the other information firm to the original image. \ourmethod is also applicable to different AR-based models, the first row is generated based on LlamaGen~\cite{sun2024autoregressive}, while the second row based on Lumina-mGPT~\cite{liu2024lumina_mgpt}
    % is able to achieve the attribute editing (left), object swapping (mid) and global style transfer (right) tasks while maintaining the other information firm to the original image.
    }
    \label{fig:teaser}
    \vspace{-1mm}
\end{center}
}]
\footnotetext[1]{ Equal contribution, † Co-corresponding authors.}
% \footnotemark[2]{Co-corresponding authors.}
\begin{abstract}
Text-to-image generation has seen groundbreaking advancements with diffusion models, enabling high-fidelity synthesis and precise image editing through cross-attention manipulation. 
Recently, autoregressive (AR) models have re-emerged as powerful alternatives, leveraging next-token generation to match diffusion models.
% while offering inherent support for localized editing. 
However, existing editing techniques designed for diffusion models fail to translate directly to AR models due to fundamental differences in structural control. 
Specifically, AR models suffer from spatial poverty of attention maps and sequential accumulation of structural errors during image editing, which disrupt object layouts and global consistency. 
In this work, we introduce Implicit Structure Locking (\ourmethod), the first training-free editing strategy for AR visual models. 
Rather than relying on explicit attention manipulation or fine-tuning, \ourmethod preserves structural blueprints by dynamically aligning self-attention patterns with reference images through the Anchor Token Matching (\ouratm) protocol.
By implicitly enforcing structural consistency in latent space, our method \ourmethod enables structure-aware editing while maintaining generative autonomy. 
Extensive experiments demonstrate that \ourmethod achieves high-quality, structure-consistent edits without additional training and is superior or comparable to conventional editing techniques. 
Our findings pioneer the way for efficient and flexible AR-based image editing, further bridging the performance gap between diffusion and autoregressive generative models.  The code will be publicly available at \url{https://github.com/hutaiHang/ATM}
\end{abstract}

\vspace{-6mm}
\section{Introduction}
\label{sec:intro}
In recent years, text-to-image (T2I) generation technologies centered on diffusion models have achieved revolutionary breakthroughs~\cite{Rombach_2022_CVPR_stablediffusion,ho2020ddpm,ho2022imagen,chen2023pixartalpha,playground-v2}. Foundational models like Stable Diffusion~\cite{podell2023sdxl,esser2024scaling} and Imagen~\cite{saharia2022photorealistic} have not only driven innovations in artistic creation~\cite{geng2024factorized_diffusion,geng2024visual_anagrams,butt2024colorpeel,gomez2024colorillusion}, but also catalyzed a rich ecosystem of downstream tasks — ranging from text-guided image editing~\cite{hertz2022prompt,tumanyan2022plug,kai2023DPL,li2023stylediffusion} and drag manipulation~\cite{Shi2023DragDiffusion:Editing,Mou2023DragonDiffusion:Models} to image-guided T2I generation~\cite{zhang2023controlnet,ye2023ip,mu2025editar,mou2024t2i_adapter,zhao2025controlvideo}. These methods generally achieve pixel-level image manipulation by exploring latent features~\cite{zhang2023controlnet,mou2024t2i_adapter} or attention weights~\cite{parmar2023zero,tumanyan2022plug} during the denoising process. 

Most recently, autoregressive models (AR) have experienced a resurgence in image generation. 
Inspired by the success of large language models (LLMs)~\cite{touvron2023llama,achiam2023chatgpt4,liu2024deepseek}, recent visual autoregressive models~\cite{xie2024show-o,wu2024janus,tian2025var,liu2024lumina_mgpt}, including LlamaGen~\cite{sun2024autoregressive} and Emu3~\cite{wang2024emu3}, treat images as discrete token sequences~\cite{esser2021taming,wang2025omnitokenizer}, reconstructing high-fidelity visuals through next-token prediction. 
These approaches match diffusion models in long-range coherence while offering unique advantages: their sequential generation mechanism inherently supports localized editing and seamless integration with multimodal language models. 
However, the potential of AR models remains underexplored for image editing tasks — their next-token generation paradigm fundamentally differs from the parallel denoising mechanism of full latent vectors in diffusion models, which prevents direct migration of existing diffusion-based editing techniques.

\begin{figure}[t]
    \centering
    \includegraphics[width=0.999\linewidth]{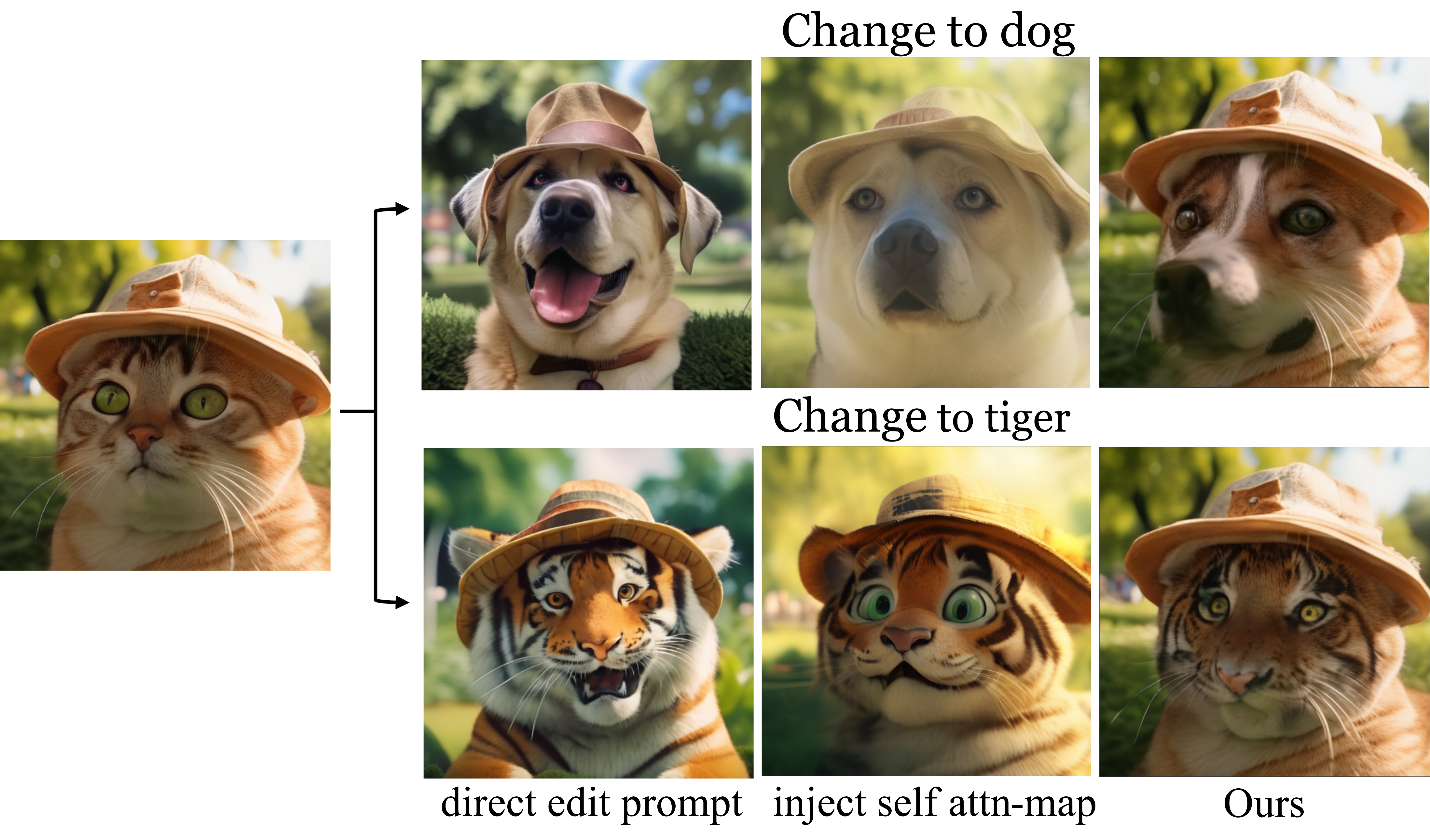}
    \vspace{-8mm}
    \caption{
    (Left) Direct modification of target tokens causes significant content distortion due to sequential structural error accumulation.
    (Mid) Naïve attention injection disrupts content coherence.
    (Right) In contrast, our implicit structure locking (\ourmethod) effectively mitigates these issues through the proposed Anchor Token Matching (\ouratm) strategy.
    }
    \vspace{-7mm}
    \label{fig:naive_edit}
\end{figure}

This challenge comes from the overlying difference in their structural control mechanisms during image generation. 
In diffusion models, text-to-image spatial correspondences are explicitly established via cross-attention maps: coarse structural information is locked during early denoising stages, allowing global consistency preservation through localized attention adjustments~\cite{hertz2022prompt,tumanyan2022plug,hu2025token}. 
In contrast, AR models follow a distinct structural generation logic: Each token prediction strictly depends on preceding sequences, with structural information not centrally determined at any single stage but progressively evolving during generation. 
This mechanism introduces two critical issues:
\textit{(1) Spatial Poverty of Attention Maps}: Text-to-image attention maps in AR models lack precise structural correspondence\footnote{Reasonably, the current token attends heavily to the previous one due to the next-token prediction mechanism in AR models.}, making them unreliable as editing anchors (as shown in Fig.~\ref{fig:attn_map}).  
\textit{(2) Sequential Accumulation of Structural Errors}: Naive modification of target tokens (e.g. changing \quotes{cat} to \quotes{dog}) induces localized shifts in latent states. These deviations propagate through autoregressive dependency chains, ultimately distorting global structures such as object poses and scene layouts (as shown in Fig.~\ref{fig:naive_edit}).  

Recent attempts~\cite{mu2025editar,li2024controlar} to mitigate these issues involve fine-tuning the model parameters in large-scale paired image editing datasets~\cite{brooks2022instructpix2pix,fu2023guiding} or introducing distillation losses. However, such approaches require substantial training data and computational resources while sacrificing zero-shot editing flexibility.
Thus, a key challenge emerges: How can we achieve structure-consistent editing in text-to-image autoregressive models in a training-free manner, leveraging a deeper understanding of their attention mechanisms?

In this work, we first conduct a systematic investigation into the structural control mechanisms inherent to autoregressive (AR) image generation. While prior research has extensively examined attention-guided editing in diffusion models~\cite{hertz2022prompt,tumanyan2022plug,kai2023DPL,parmar2023zero}, the relationship between attention maps and structural layout in AR frameworks remains unexplored.
Existing diffusion-based approaches~\cite{hertz2022prompt,tumanyan2022plug} suggest that transplanting reference attention maps can effectively enforce structural consistency. 
However, our experiments reveal a fundamental limitation of this approach when applied to AR models:
injecting external attention maps disrupts the intrinsic attention dynamics of AR models, leading to coherence loss with the semantic context in the generated image (as illustrated in Fig.~\ref{fig:naive_edit}-mid, The layout structure is kept but the content is distorted.).
This disruption manifests itself as blurred textures, distorted object proportions, and inconsistent lighting, artifacts arising from the inherent incompatibility between parallel attention injection and the sequential dependencies of AR models.
% \KW{implicit approximation of attention maps. We also makes the non-edit region unchanged.}
To address this challenge, we propose Implicit Structure Locking (\ourmethod) as the first zero-shot training-free editing strategy for AR visual generative models.
At the core of our approach is the Anchor Token Matching (\ouratm) strategy. Instead of brute-force attention map transplantation, we selectively match tokens during autoregressive decoding by identifying those whose hidden representations exhibit maximum similarity to anchor tokens from the original sequence. 
This process induces implicit attention alignment, enabling the model to naturally compute attention maps that preserve structural coherence while adapting to local semantic edits (e.g., transitioning fur texture from \quotes{cat} to \quotes{dog}). Note that attention consistency emerges as a by-product of \ourmethod rather than from explicit constraints.
% , ensuring seamless and structure-consistent modifications.
And that \ourmethod simultaneously achieves two critical objectives: 
\textit{(1) Preserving Structural Blueprints:} the self-attention maps of the edited sequence maintain structural consistency with those of the reference image. 
\textit{(2) Maintaining Generative Autonomy:} allowing the model to dynamically adjust attention patterns to accommodate the semantic requirements of the edited content, while the non-edit regions are unchanged.

% — a paradigm we term "structure-through-geometry" inheritance. 
% \KW{We proposed 2 objectives and we then solve the objectives by our proposal methods.}

Through extensive comparisons with existing diffusion-based and AR model image editing approaches on the widely used PIE-Bench~\cite{direct_inversion_2023} dataset, our proposed method, \ourmethod, achieves user satisfactory performance in text-guided image editing. In summary, our key contributions include:
\begin{itemize}[leftmargin=*]
    \item We conduct an in-depth study of attention mechanisms in autoregressive (AR) image generation, revealing the limitations of existing diffusion-based structural control methods when applied to AR models. 
    % % (\KW{It is our byproduct to align the attention, not the reason to achieve editing.})
    % \item Based on this finding, we propose the \textit{first} training-free editing method, Implicit Structure Locking (\ourmethod), that approximates structural layouts by preserving self-attention coherence during autoregressive decoding, overcoming spatial inconsistencies in text-guided editing.
    % \item To maintain attention coherence, we introduce a token-matching mechanism, Anchor Token Matching (\ouratm), that implicitly preserves key structural elements by identifying anchor tokens in latent space, ensuring semantic consistency without disrupting generative autonomy.
    % \item Extensive experiments on the PIE-Bench~\cite{direct_inversion_2023} demonstrate that \ourmethod significantly maintains the structural consistency and visual fidelity in AR-based image editing.
    
    % \item We propose \ourmethod, the \textit{first} training-free editing framework for AR models that achieves structure-preserving generation through latent geometry alignment. Rather than explicitly enforcing attention map consistency, our method implicitly maintains structural fidelity by preserving the hidden state trajectory's continuity during decoding.  
    \item Based on this finding, we propose the \textit{first} training-free editing method, Implicit Structure Locking (\ourmethod), that approximates structural layouts by implicitly matching attention pattern during autoregressive decoding, overcoming spatial inconsistencies in text-guided editing.
    
    \item We introduce a token-matching mechanism, Anchor Token Matching (\ouratm), which implicitly preserves key structural elements by identifying anchor tokens in latent space while allowing natural emergence of attention coherence as a byproduct, ensuring semantic consistency without disrupting generative autonomy.
    
    \item Extensive experiments on the PIE-Bench~\cite{direct_inversion_2023} demonstrate that \ourmethod significantly maintains the structural consistency and visual fidelity in AR-based image editing.  
\end{itemize}

\section{Related Work}
\label{sec:related_work}

% \KW{This subsection not necessary. TO REMOVE!\{ }
% \minisection{Diffusion-based Image Generating.}
% Denoising Diffusion Probabilistic Models (DDPMs)~\cite{ho2020ddpm} revolutionized image synthesis by reformulating generation as an iterative denoising process. Stable Diffusion~\cite{rombach2022high, podell2023sdxl} advanced this paradigm through latent space diffusion, reducing computational costs while preserving high fidelity. Its U-Net architecture~\cite{ronneberger2015unet} integrates text conditioning via cross-attention layers~\cite{vaswani2017attention}, establishing the foundation for text-to-image synthesis. Recent innovations challenge this convention: The Diffusion Transformer (DiT)~\cite{Peebles2022DiT,bao2023all} replaces U-Net with pure transformer blocks, demonstrating scaling law advantages; MM-DiT architecture~\cite{esser2024scaling} enhances multimodal alignment through full attention mechanisms; Flow-Matching~\cite{esser2024scaling,liu2022flow, lipman2022flow} accelerates inference via continuous-time flow matching. These advances drive diffusion models toward computational efficiency and architectural unification. Notably, the autoregressive paradigm has demonstrated breakthrough progress in image generation, driven by advancements in language modeling~\cite{achiam2023chatgpt4,touvron2023llama}.
% \KW{TO REMOVE above! ===== \} }

\minisection{Autoregressive Image Generation.}
Inspired by the sequence prediction paradigm in LLMs~\cite{achiam2023chatgpt4,liu2024deepseek}, autoregressive (AR) models reformulate image generation as an image token sequence prediction task. 
The pioneering PixelCNN~\cite{van2016conditional} achieved image synthesis through pixel-wise conditional probability modeling, yet its limited receptive-field constrained global coherence. 
Subsequent works in discrete representation learning addressed this bottleneck: VQ-VAE~\cite{razavi2019generating} and VQGAN~\cite{esser2021taming} established learnable discrete codebooks that compress images into token sequences, laying the foundation for scalable AR modeling. MaskGIT~\cite{chang2022maskgit} innovatively introduced masked prediction mechanisms, enabling parallel decoding through bidirectional contextual modeling while preserving autoregressive properties. 
% VAR~\cite{tian2025var} innovatively shifted next-token prediction to next-scale prediction based on~\cite{lee2022autoregressive_residual}, which predicts all tokens within a specific scale aids in maintaining internal consistency within the image.

With the progress of LLMs, researchers have explored cross-modal extensions—LlamaGen~\cite{sun2024autoregressive} adapted LLaMA architectures~\cite{touvron2023llama} for visual token modeling, achieving visual results comparable to diffusion models~\cite{Rombach_2022_CVPR_stablediffusion,deepfloyd,chen2023pixartalpha}, while Emu3~\cite{wang2024emu3} constructed a unified autoregressive space for multimodal joint training. Notably, recent work~\cite{chen2025janus,xie2024show-o} further integrates visual understanding and generation within a single autoregressive framework, demonstrating task generalization potential. 
However, existing AR-based approaches encounter fundamental challenges in controllable image editing. Their sequential generation process inherently accumulates errors during localized modifications, leading to discrepancies that conflict with the strict spatial consistency required for precise image editing tasks.

\minisection{Text-guided Image Editing.} %, 
Text-guided image editing aims to modify image content based on semantic prompts while preserving irrelevant regions. Traditional GAN-based methods~\cite{goodfellow2014gan, liu2023survey, patashnik2021styleclip, patashnik2021styleclip} achieve local editing by optimizing in the GAN latent space with CLIP guidance~\cite{jiang2025cmsl, peng2023unsupervised}, but their performance is constrained by the capacity of pre-trained GANs. 
Recent diffusion-based approaches have become mainstream, yet they often require complex inversion and latent optimization processes to balance fidelity and controllability. 
% For instance, SDEdit~\cite{meng2022sdedit} edits images by adding noise and denoising with target text prompts, but struggles to maintain fidelity with the original image. 
For example, optimization-based inversion methods~\cite{mokady2022null,kai2023DPL,li2023stylediffusion,han2023ProxNPI} refine latent noise for accurate reconstruction and manipulate cross-attention maps (e.g., Prompt-to-Prompt~\cite{hertz2022prompt}) to preserve structure.
InstructPix2Pix~\cite{brooks2022instructpix2pix} bypasses inversion by training on paired edit data, but relies on Prompt-to-Prompt~\cite{hertz2022prompt} to generate large-scale image-instruction pairs. To reduce data dependency, some work~\cite{cao2024instruction,chen2024unireal} extracts consecutive video frames as editing samples to simulate real-world editing dynamics. Recent studie ~\cite{mu2025editar} adapt the InstructPix2Pix framework to autoregressive models through fine-tuning on paired datasets. 

However, most existing editing methods remain fragmented due to their underlying architectures—while diffusion models dominate the landscape, AR models remain underexplored, primarily due to the absence of a zero-shot editing framework.
A few prior works~\cite{mu2025editar,li2024controlar} have attempted to address this limitation through computationally expensive fine-tuning on large-scale paired image editing datasets, often at the cost of zero-shot flexibility.
In contrast, our approach unifies spatial control and semantic editing within the AR paradigm, eliminating the need for optimization, or additional model components.
\section{Method}
\label{sec:method}
\begin{figure*}
    \centering
    \includegraphics[width=0.999\linewidth]{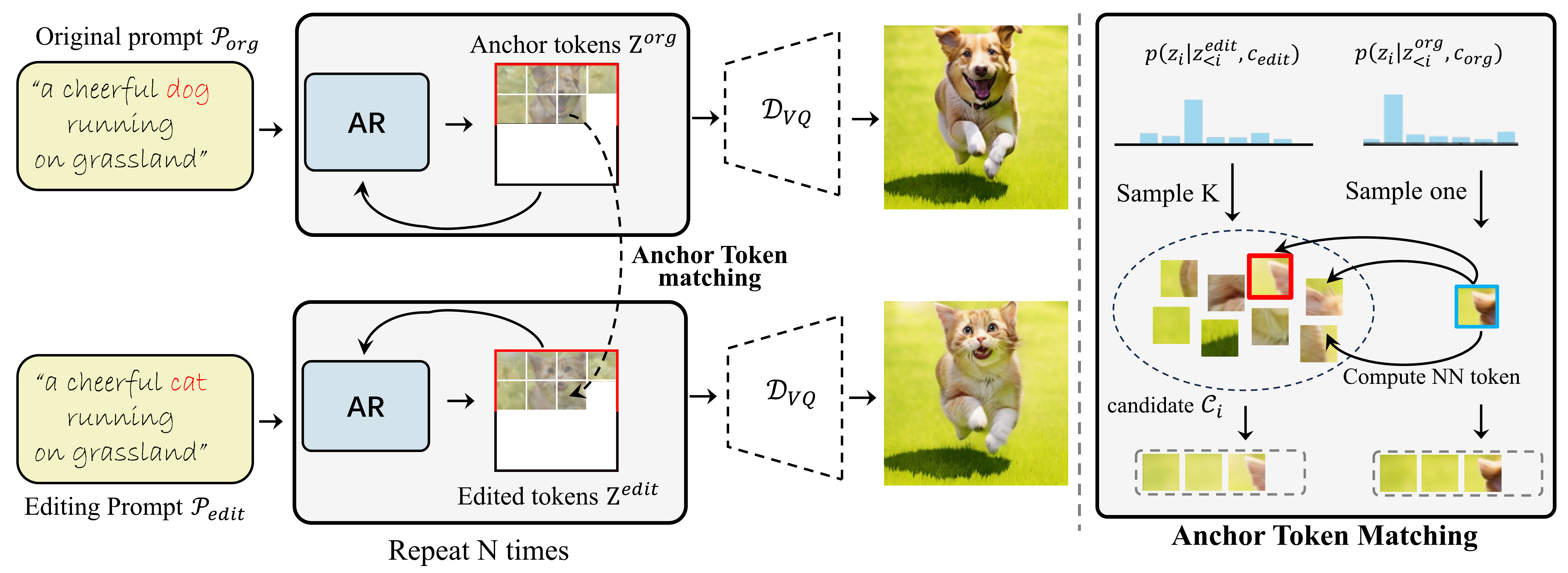}
    \vspace{-7mm}
    \caption{Our method, \ourmethod, achieves implicit structural locking through Anchor Token Matching (\ouratm). Rather than relying on direct attention injection—which often introduces distortions in AR-based visual generation—our approach selects the optimal editing token by identifying the candidate with the smallest distance to the reference token among  $K$  candidates.
    }
    \vspace{-5mm}
    \label{fig:method}
\end{figure*}

This work aims to achieve zero-shot text-guided image editing using text-to-image autoregressive (AR) models~\cite{sun2024autoregressive} by systematically investigating the key factors that govern image structure in AR-based generation.
We begin by briefly reviewing the paradigm of text-to-image AR models in Section~\ref{sec:pre}, which serves as the foundational framework for our approach. Next, in Section~\ref{sec:ana}, we establish our motivation through a series of analytical experiments that highlight the structure preservation challenges in AR-based editing. Finally, we introduce our method, Implicit Structure Locking (\ourmethod), in Section~\ref{sec:IGL}, detailing its design and effectiveness. An overview of \ourmethod is illustrated in Fig.~\ref{fig:method}.

\subsection{Preliminary}
\label{sec:pre}
The AR visual generation model LlamaGen~\cite{sun2024autoregressive} synthesizes images from text by sequentially predicting image tokens. Its architecture consists of two key components working in synergy: a VQ-Autoencoder~\cite{razavi2019generating,esser2021taming} that converts images into discrete token sequences and an autoregressive transformer $f_{\theta}$ that learns the joint distribution of these image tokens. 
Given an input image $x \in \mathcal{R}^{H \times W \times 3}$, the feature encoder $\mathcal{E}_{VQ}$ first maps it to a latent representation $z^e \in \mathcal{R}^{h \times w \times d}$, where $d$ denotes the feature dimension. Through nearest-neighbor quantization, each spatial feature vector $z^e_{i,j}$ is projected to a codebook prototype $z_{i,j}^q \in \mathcal{V}$, generating a discrete token sequence $\mathbf{Z} = \{z_1, ..., z_{h \times w}\}$, where $\mathcal{V}$ is the learned codebook vector set. 
To enable text-conditioned generation, LlamaGen~\cite{sun2024autoregressive} integrates a pre-trained T5~\cite{raffel2020exploring} text encoder $ \tau_\xi $ that maps the text prompt $ \mathcal{P} $ to a sequence of embeddings $ c = \tau_\xi(\mathcal{P})$. 
The text embeddings are projected into the transformer input space through a linear layer and prepended to the image token sequence. 
Following that, the AR model $f_\theta$ autoregressively predicts the joint distribution over image tokens conditioned on the text: 
\vspace{-1mm}
\begin{equation}
P(z|c) = \prod_{i=1}^{N} P(z_i|z_{<i}, c)
\vspace{-1mm}
\end{equation}
where $ N = h \times w $. During generation, the concatenated sequence $ [c; {z}_{<i}] $ is processed through stacked transformer layers with causal masking to enforce autoregressive constraints. At each layer $ l $, the self-attention mechanism is:  
% \vspace{-2mm}
\begin{equation}
\mathbf{A}_l = \text{Softmax}\left(\frac{\mathbf{Q}_l \mathbf{K}_l^\top}{\sqrt{d_k}}\right)
\vspace{-2mm}
\end{equation}
where $ \mathbf{H}_l $ denotes hidden states at layer $ l $, $ W_l^Q, W_l^K, W_l^V $ are projection matrices and $\mathbf{Q}_l = \mathbf{H}_l W_l^Q, \mathbf{K}_l = \mathbf{H}_l W_l^K, \mathbf{V}_l = \mathbf{H}_l W_l^V$. The final layer applies an MLP followed by softmax to predict the next token distribution over $ \mathcal{V} $. The generated token sequence $ \mathbf{s} $ is ultimately decoded to RGB image space via the decoder $ \mathcal{D}_{\text{VQ}} $, completing the text-to-image synthesis pipeline.

\subsection{Structural Information Analysis}
\label{sec:ana}
% \Th{to complete the fig and table}
\begin{figure}[t]
    \centering
    \includegraphics[width=0.999\linewidth]{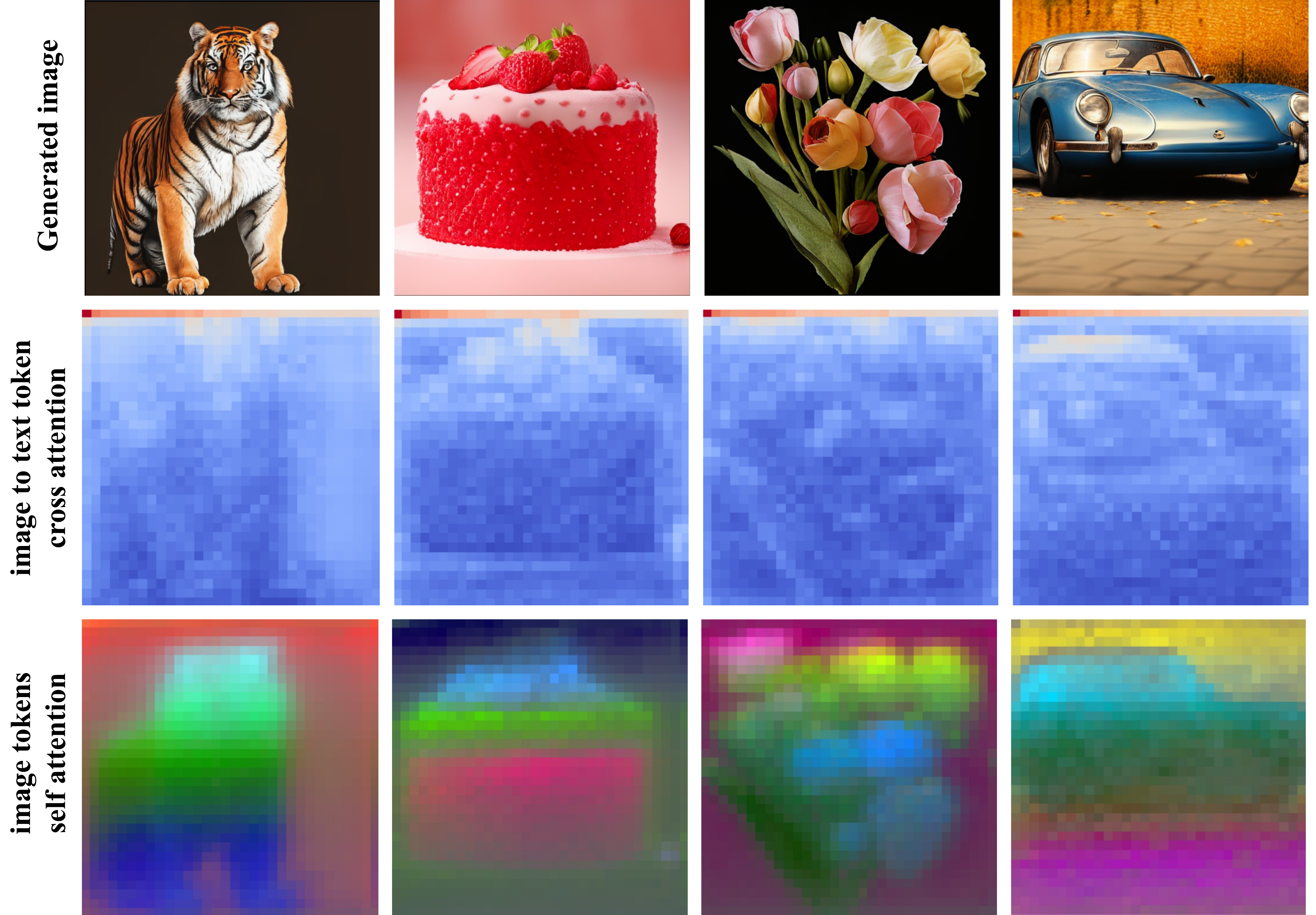}
    \vspace{-6mm}
    \caption{The cross-attention (2nd row) and self-attention (3rd row) visualization after generating the last token with LlamaGen model. We observe that cross-attention from image to text tokens contains minimal structural information, whereas self-attention maps exhibit stronger semantic alignment with the structural layout.
    }
    \vspace{-5mm}
    \label{fig:attn_map}
\end{figure}
To uncover the intrinsic structural control mechanisms of AR visual generation models, we conducted a series of systematic experimental analyses focused on their \textit{attention dynamics} and \textit{sequential sensitivity}. Our observations reveal a crucial insight: although the cross-attention maps connecting text to image tokens lack significant spatial information\footnote{In AR models, cross-attention differs from diffusion models by sharing QKV projections across text and image tokens, unlike diffusion models that use separate mappings. More details see in supplementary}, \textit{the self-attention maps among image tokens exhibit rich structural information.}
% spatial and semantic organizational structures.

As illustrated in Fig.~\ref{fig:attn_map}, we applied the Principal Component Analysis (PCA) decomposition to the self-attention matrix \( A \in \mathcal{R}^{(h \times w) \times (h \times w)} \) to a three-dimensional space. 
The visualization results indicate that semantically similar tokens tend to exhibit coherent attention patterns, suggesting that spatial structures are formed through image token self-organization, rather than relying solely on explicit text prompt guidance.
However, during the injecting the attention maps of reference images into the target generation process as a naive way to preserve the structural consistency, we observed significant artifacts and global distortions (as shown in Fig.~\ref{fig:naive_edit}). 
% Quantitative analysis revealed that the Fréchet Inception Distance (FID) increased by 23\% compared to the baseline generation,
Which we attribute to the contextual mismatch between the reference attention maps and the latent dynamics of the target sequence. 

\begin{figure}[t]
    \centering
    \includegraphics[width=0.999\linewidth]{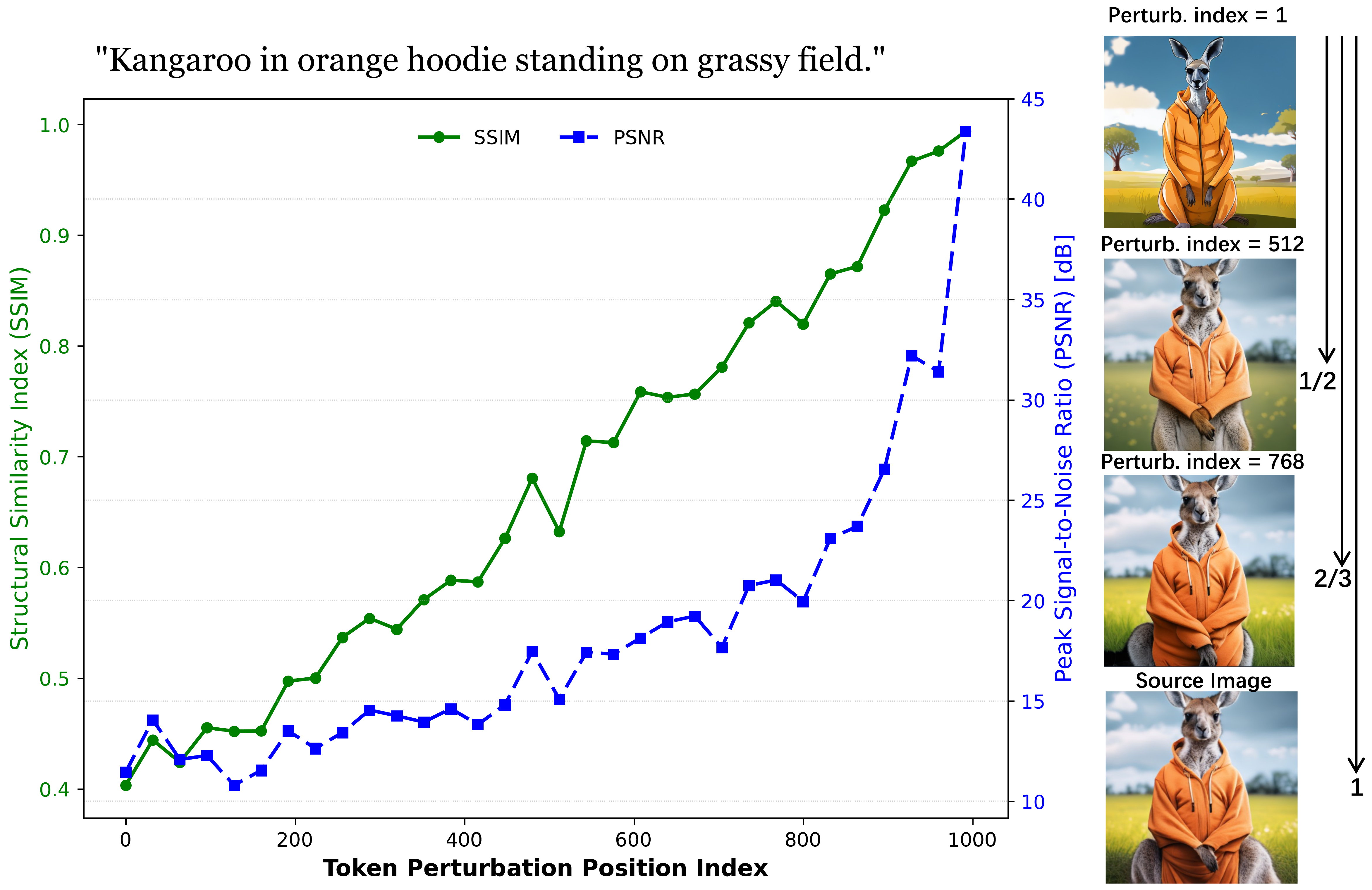}
    \vspace{-7mm}
    \caption{
    Perturbing image tokens at different stages of AR generation results in varying changes in image quality metrics (SSIM and PSNR), as shown in the left curves. Early-stage perturbations primarily affect global structural geometry, while later-stage perturbations influence only high-frequency details, as evident from the generated results on the right.
    }
    \vspace{-6mm}
    \label{fig:pert}
\end{figure}

As shown in Fig.~\ref{fig:pert}, Further studies on \textit{sequential sensitivity} reveal that perturbing the first 20\% of tokens in the generation sequence leads to a significant decrease in the Structural Similarity Index (SSIM), with a change of \( 0.56 \pm 0.02 \).
This impact is notably higher than the effect of perturbations occurring later 20\% in the sequence, where the SSIM change is \( \Delta \text{SSIM} = 0.08 \pm 0.05 \).
Moreover, distortions caused by late-stage perturbations are predominantly concentrated in high-frequency detail areas, such as fine textures and edges.
This observed phenomenon of progressive structural solidification aligns with theoretical expectations: early tokens, owing to the causal attention mechanism in transformers, interact with all subsequent positions and thus play a critical role in shaping the global image structure. In contrast, later tokens are more constrained by the local contexts of previously generated tokens, leading to a lesser influence on the overall structure. This highlights the importance of early-stage tokens in defining the global image structure during autoregressive image generation.

\subsection{Implicit Structure Locking (\ourmethod)}
\label{sec:IGL}

Building on our observations from the previous section, we find that directly injecting attention maps leads to significant artifacts and distortions, disrupting the coherence of the generated image. To address this issue and ensure structural consistency during image editing, we propose an adaptive decoding framework that leverages anchor token matching (\ouratm). Instead of explicit attention transplantation, our \ourmethod implicitly locks the structure in the latent space.

\minisection{Anchor Token Matching with Dynamic Windows.}
% Given a reference image $ x^{\text{org}} $, we first obtain its discrete token sequence $ \mathbf{Z}^{\text{org}} = [z_1^{\text{org}}, ..., z_N^{\text{org}}] $ via a VQ encoder $ \mathcal{E}_{\text{VQ}} $. 
Given the original prompt $\mathcal{P}_{org}$ and the editing prompt $\mathcal{P}_{edit}$, the AR model samples the next token based on the predicted distribution $p(z_i|z_{<i},c)$. For the $\mathcal{P}_{org}$, it samples one token $z_i^{org}$ from distribution $p(z_i|z_{<i}, c_{org})$, which serves as the anchor token at the corresponding position when generating the edited image. Our dynamic anchor token matching (\ouratm) framework implicitly aligns structures while maintaining editing flexibility. At step $ i $ of generating the edited sequence $ \mathbf{Z}^{\text{edit}} $, we sample $ K $ candidate tokens $ \mathcal{C}_i = \{z_i^{(1)}, ..., z_i^{(K)}\} \subset \mathbb{R}^d $ from the conditional distribution $ p(z_i | z_{<i}^{\text{edit}}, c_{edit}) $. We compute the latent space Euclidean distance between each candidate $z_i^{(k)}$ and the reference anchor $z_i^{\text{ref}}$ as:  
\begin{equation}   
s^{(k)} = \|z^{(k)}_i - z_i^{\text{org}}\|^2_2 
\end{equation}

We then select the matching candidate with the minimal distance as the output, following a process akin to nearest-neighbor (NN) computation.
This \ouratm strategy ensures local alignment between the hidden state trajectories of the edited and reference sequences, guiding the self-attention mechanism to generate structurally consistent attention maps. 
By leveraging implicit structural constraints rather than directly injecting attention maps, our approach \ourmethod mitigates semantic conflicts and preserves contextual coherence, enabling more stable and structure-aware image editing.

To accommodate the varying structural constraints across diverse generation stages, we further introduce a \textit{dynamic window} mechanism. 
The adaptive filtering window $ \mathcal{W}_i \subseteq \mathcal{C}_i $ adjusts its size proportionally to decoding progress as:  
\vspace{-1mm}
\begin{equation}
|\mathcal{W}_i| = \lfloor K \cdot (1 - \alpha \cdot \frac{i}{N}) \rfloor, \quad \alpha \in [0,1]  
% \vspace{-1mm}
\end{equation}
where we set $\alpha = 0.6$ by default. At initialization (\(i=0\)), the window retains \(100\%\) of candidates (\(|\mathcal{W}_i|=K\)) to enforce strict structure alignment for foundational structures. As generation progresses, the window shrinks linearly, reaching \(70\%\) capacity (\(|\mathcal{W}_i| \approx 0.7K\)) at \(i=0.5N\) and \(40\%\) (\(|\mathcal{W}_i| \approx 0.4K\)) by completion (\(i=N\)). This design guarantees continuous adaptation, as early phases prioritize structural fidelity through broad candidate pools, while later stages progressively emphasize contextual coherence via tighter constraints.
% During initial generation ($ i < 0.2N $), we set $ \alpha = 0.3 $ to retain $ \approx 70\% $ candidates ($ |\mathcal{W}_i| \approx 0.7K $), leveraging early tokens' dominance in global structure formation. In later stages ($ i > 0.8N $), $ \alpha = 0.9 $ reduces the window to $ \approx 10\% $ candidates ($ |\mathcal{W}_i| \approx 0.1K $), unleashing model creativity for detail refinement.
The final editing token is noted as:
\vspace{-0.5mm}
\begin{equation}
z_i^{\text{edit}} = \arg\min_{k \in \mathcal{W}_i} s^{(k)}
\vspace{-1mm}
\end{equation}
By this means, our method \ourmethod achieves progressive latent space alignment, effectively mitigating pattern mismatch through implicit structural guidance. 

\minisection{Adaptive Constraint Relaxation (\ouracr).}
To balance \textit{structural blueprints} and \textit{generative autonomy}, we propose an adaptive threshold-based autonomy-preservation scheme. 
A similarity threshold $ \tau $ works as a constraint regulator:  
\begin{equation}
z_i^{\text{edit}} = \begin{cases}  
\arg\min s^{(k)} & \text{if } \min s^{(k)} \leq \tau \\  
\arg\max p(z_i | z_{<i}^{\text{edit}},c_{edit}) & \text{otherwise}  
\end{cases}  
\end{equation}

% where $ \tau $ linearly decays from 0.8 (strict constraints) to 0.4 (relaxed constraints) during decoding. This ensures structural consistency in structurally critical phases ($ i < 0.3N $) while allowing maximum likelihood dominance in detail refinement ($ i > 0.7N $).
The mechanism incorporates two safeguards: (1) candidate window pre-filtering for AR generation quality and (2) dynamic thresholds to prevent over-constraint errors.
% We set $\tau=1.0$ as the default hyperparameter. 
Moreover, users can adjust $\tau$ to balance between text-to-image generation diversity and similarity to the original input image based on their specific requirements.
A larger $\tau$ preserves greater similarity to the original image, while a smaller $\tau$ allows for increased generation diversity.

% \minisection{Cooperative Attention-Guided Decoding (Optional).}

% \minisection{Attention Fusion (\ourhaf).}
% % \KW{Only userful for global editing -  style transfer.}
% % \KW{Taihang write why it is only useful for global case.}
% To improve structural consistency, particularly in global style transfer tasks, we introduce an attention fusion mechanism (\ourhaf).
% As shown in Section~\ref{sec:ana}, direct injection of reference attention maps deteriorates image quality. However, when combined with the anchor token matching (\ouratm), our \ourmethod enhances structural coherence while preserving fidelity.
% More specifically, in shallow layers ($ l \in \{1, ..., L/4\} $) responsible for global structure modeling, we linearly fuse reference and generated editing attention maps:  
% $$
% \mathbf{A}_i^{\text{hybrid}} = \lambda \cdot \mathbf{A}_i^{\text{ref}} + (1-\lambda) \cdot \mathbf{A}_i^{\text{edit}}  
% $$
% The blending coefficient $ \lambda $ linearly decays with depth ($ \lambda = 0.7 - 0.3 \cdot \frac{l}{L/4} $), allowing lower layers to inherit reference structural patterns while upper layers transition to autonomous generation. 
% \KW{Consider to move to supp as it is changing the attention maps, not well aligning with our ATM proposal.}

Finally, our \ourmethod framework is primarily built on the Anchor Token Matching (\ouratm) strategy, enhanced by additional \textit{dynamic windows} and \ouracr techniques.
These modules work in synergy to achieve training-free text-guided image editing within autoregressive (AR) generative models.
% —an unexplored direction in prior research. We introduce this framework as the first attempt to bridge this gap.
% \Th{to complete the fig and table}

% demonstrating its effectiveness through comprehensive evaluations. 
% In the following section, we conduct ablation studies to assess the individual contributions of each component to the overall performance.

\begin{figure*}[t!]
    \centering
    \vspace{-1mm}
    \includegraphics[width=0.965\linewidth]{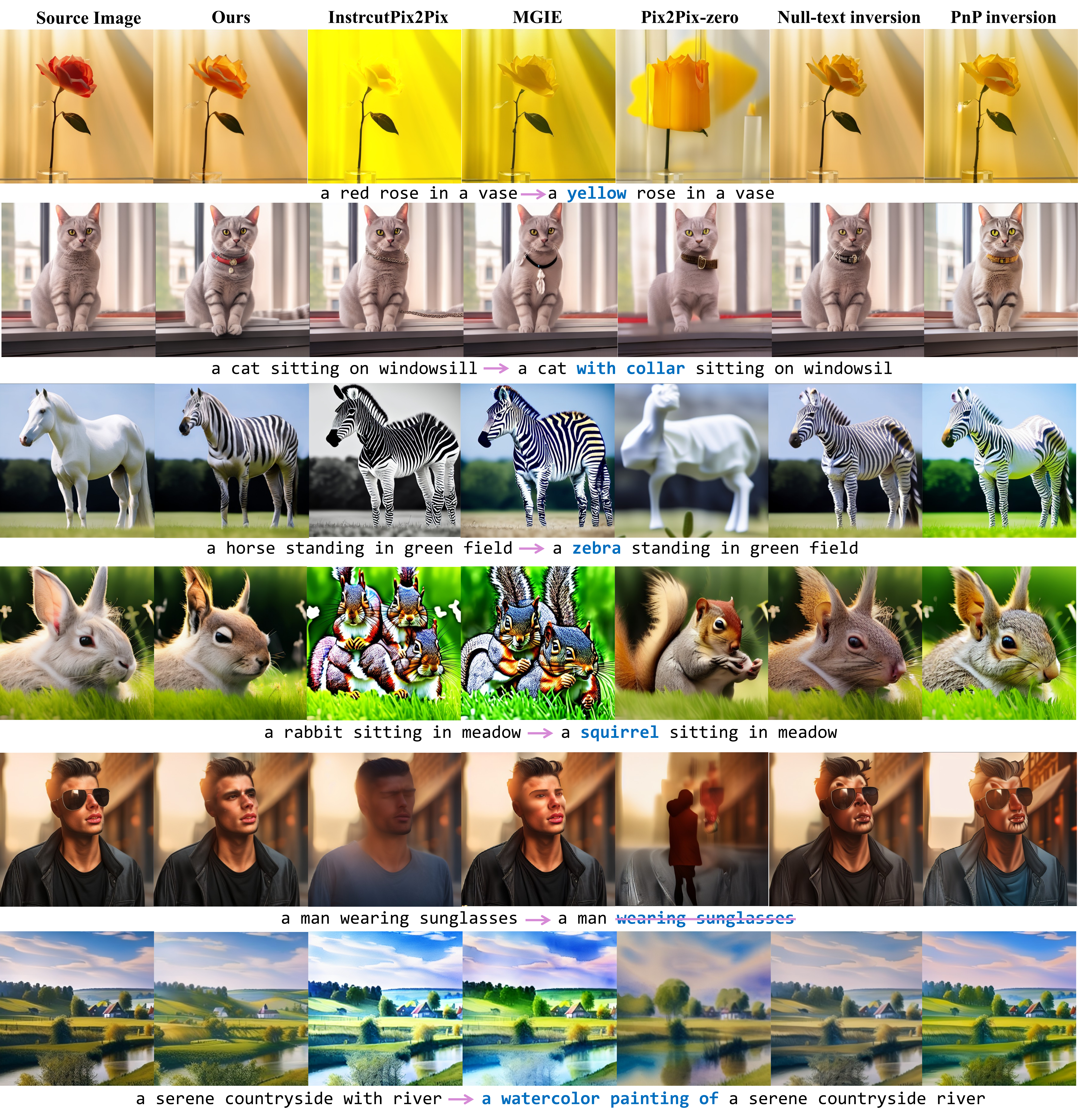}
    \vspace{-4mm}
    \caption{Qualitative comparison with various text-guided image editing methods.}
    \vspace{-5mm}
    \label{fig:qualitative_comp}
\end{figure*}

\begin{table*}[t!]
\centering
\resizebox{0.957\textwidth}{!}{
\begin{tabular}{c|c|c|c|c|c|c|c|c}
\toprule[2pt]
\multicolumn{1}{c|}{} & \multicolumn{1}{c|}{} & \multicolumn{1}{c|}{\textbf{Structure}} & \multicolumn{4}{c|}{\textbf{Background Preservation}} & \multicolumn{2}{c}{\textbf{CLIP Similarity}} \\ \cline{3-9} 
\multicolumn{1}{c|}{\multirow{-2}{*}{\textbf{Method}}} & {\multirow{-2}{*}{\textbf{\begin{tabular}[c]{@{}c@{}}T2I\\Model\end{tabular}}}} & \multicolumn{1}{c|}{\textbf{Distance} $\downarrow$} & \multicolumn{1}{c|}{\textbf{PSNR} $\uparrow$} & \multicolumn{1}{c|}{\textbf{LPIPS} $\downarrow$} & \multicolumn{1}{c|}{\textbf{MSE} $\downarrow$} & \multicolumn{1}{c|}{\textbf{SSIM} $\uparrow$} & \multicolumn{1}{c}{\textbf{Whole} $\uparrow$} & \multicolumn{1}{c}{\textbf{Edited} $\uparrow$}\\ \cline {1-9}
Prompt-to-Prompt~\cite{hertz2022prompt} & SD1.4 & 88.46 & 16.80 & 270.38 & 241.89 & 69.93 & 26.70 & 21.43 \\
Null-text Inversion~\cite{mokady2022null}& SD1.4 &18.42 & 25.68 & 77.70 & 42.92 & 85.71 & 24.55 & 20.73   \\
Pix2pix-zero~\cite{parmar2023zero} & SD1.4 &59.43 & 19.71 & 193.44 & 147.19 & 76.48  & 23.56 & 19.76 \\
MasaCtrl~\cite{cao2023masactrl} & SD1.4 & 34.20 & 21.59 & 124.35 & 83.60 & 81.31 & 22.90 & 18.52 \\
% DDIM+PnP & SD1.5 &  &  &  &  &  &  &  \\
PnPInversion~\cite{direct_inversion_2023} & SD1.5 & 24.81 & 22.16 & 114.15 & 74.07 & 81.81 & 25.56 & 21.50  \\
\bottomrule
InstructPix2Pix~\cite{brooks2022instructpix2pix} & SD1.5 & 67.49 & 19.69 & 164.27 & 235.62 & 76.98 & 23.37  &  20.48 \\
MGIE~\cite{fu2023guiding} & SD1.5 & 53.46 & 20.62 & 131.13 & 205.09 & 79.55 &22.67 & 19.58  \\
% Ours - Vanilla & LlamaGen &  &  &  &  &  &  &  \\
\bottomrule
\textit{NPM} & LlamaGen &113.95  &12.14  &377.84  &725.98  &53.67  &\textbf{24.71} &21.28   \\
PnP-AR & LlamaGen &103.94  & 13.20 &328.49  &600.20  &58.25  & 23.56&20.65   \\
\ourmethod (Ours) & LlamaGen &\textbf{31.79}  &\textbf{19.75}  &\textbf{136.21} &\textbf{161.17} &\textbf{76.71} & \underline{24.19}  & \textbf{21.33}
\\ \bottomrule[2pt]
\end{tabular}%
}
\vspace{-3mm}
\caption{We compare mainly with AR based methods. We should highlight our method in some way. 
AR-based methods with the best and second-best numbers are marked with \textbf{bold} and \underline{underlined} respectively.
}
\vspace{-4mm}
\label{tab:image-editing}
\end{table*}

\section{Experiments}
\label{sec:experiments}
\subsection{Experimental Setups}
\minisection{Benchmarks.}
We build our method \ourmethod on LlamaGen~\cite{sun2024autoregressive} to generate images at a resolution of 512$\times$512.
By default, we set $K=150,\tau = 1.0$ as our hyperparameters.
To establish a rigorous evaluation framework for training-free AR image editing, we designed a generation-to-editing pipeline that aligns with our method's operational paradigm. Given the intrinsic requirement of our approach to process AR-generated images, we evaluated our method on a curated subset of the PIE-Bench dataset~\cite{ju2023direct}, the standard benchmark for image editing. While existing methods typically support all 10 editing types in PIE-Bench, we focused on \textit{5 fundamental categories} compatible with our AR training-free paradigm: object replacement, object addition, object removal, style transfer, and attribute modification. Each case in the curated subset of the PIE-Bench~\cite{direct_inversion_2023} includes paired original and editing prompts, we first used LlamaGen to generate images based on original prompts, and then applied \ourmethod to edit these images according to editing prompts.

\minisection{Metrics.} To comprehensively evaluate our method, we employed three core metrics: (1) structural consistency between original and edited images through Structure Distance~\cite{tumanyan2022splicing}; (2) background preservation quantified by PSNR, LPIPS~\cite{zhang2018unreasonable}, MSE, and SSIM~\cite{wang2003ssim} between background regions (using foreground masks generated via Grounded-SAM~\cite{ren2024grounded}); and (3) semantic alignment measured by CLIP Score~\cite{radford2021clip} for the whole image and regions in the editing mask. This multi-aspect evaluation framework ensures rigorous validation of both structural integrity and semantic fidelity in training-free AR image editing.

% \minisection{Implementation Details.}
% We default to using $K=150$, $\tau = 1.0$, with LlamaGen-XL~\cite{sun2024autoregressive} as our base model, generating all images at a resolution of 512x512.

\minisection{Comparison Methods.}
Following ~\cite{mu2025editar}, we compared the current diffusion-based text-driven image editing methods, which can be broadly categorized into two paradigms: inversion-based and inversion-free methods. The inversion-based techniques, including Prompt-to-Prompt~\cite{hertz2022prompt}, Null-text Inversion~\cite{mokady2022null}, PnPInversion~\cite{ju2023direct}, Pix2Pix-Zero, and MasaCtrl~\cite{cao2023masactrl}, typically rely on optimizing an inverted latent representation of the input image to maintain structural coherence during editing.
% , often through mechanisms like cross-attention manipulation, feature injection, or spatial-aware regularization. 
In contrast, inversion-free methods such as InstructPix2Pix~\cite{brooks2022instructpix2pix} and MGIE~\cite{fu2023guiding} bypass explicit latent inversion by leveraging alternative strategies, 
% InstructPix2Pix employs instruction-tuned diffusion models and MGIE integrates multimodal visual-textual prompts via hybrid transformers to directly steer the editing process without iterative optimization. 

\subsection{Experimental Results}

\minisection{Qualitative Comparison}.
Fig.~\ref{fig:qualitative_comp} presents a qualitative comparison between our method and other editing approaches across various editing tasks. Instruction-based methods, such as InstructPix2Pix~\cite{brooks2022instructpix2pix} and MGIE~\cite{fu2023guiding}, perform well in tasks involving object addition and global style transfer—for example, in the second row (cat→cat with collar) and the last row (photo→watercolor painting). However, these methods struggle with object replacement tasks, often introducing unintended global background changes. For instance, in the third row (horse→zebra), the transformation alters both the grass texture and background color. Similarly, in the first row (rose→yellow rose), the background color is noticeably affected. Furthermore, when these methods fail, they can introduce severe visual artifacts, as seen in the fourth row (rabbit→squirrel), where the edit results in unnatural image saturation and completely distorted content.
In contrast, inversion-based approaches such as Null-text Inversion~\cite{mokady2022null} and PnP Inversion~\cite{direct_inversion_2023} offer a better balance between background preservation and edit alignment. However, their reliance on cross-attention maps imposes a fundamental limitation: they struggle with object removal tasks. This is evident in the fifth row (man wearing sunglasses→man), where both methods fail.
% neither method successfully removes the sunglasses.

\ourmethod demonstrates strong potential across all five editing types: object addition, object removal, object replacement, attribute modification, and style transfer. It effectively preserves structural consistency while ensuring localized and precise modifications, making it a more versatile solution for text-driven image editing.

\minisection{Quantitative Comparison}.
Table~\ref{tab:image-editing} presents a comprehensive comparison between \ourmethod and diffusion-based editing approaches across multiple critical metrics. Unlike prior works predominantly reliant on Stable Diffusion~\cite{Rombach_2022_CVPR_stablediffusion}, \ourmethod represents the first exploration of training-free structured image editing based on autoregressive (AR) models. While fundamental differences in base model capabilities inherently introduce new challenges, our results demonstrate competitive performance across all evaluation dimensions.

We also implement two simple baselines: Naive Prompt Modification (\textit{NPM}), which directly modifies target words in the prompt, and PnP-AR~\cite{hertz2022prompt}, which replaces attention maps generated from the original image during editing. As shown in Table~\ref{tab:image-editing}, these baselines struggle to balance structural consistency with background preservation, exhibiting significantly higher Structure Distance scores (113.95 and 103.94, respectively) and inferior background fidelity. 
% This observation confirms that directly porting attention map-based editing schemes from diffusion frameworks to AR architectures fails to achieve controlled, localized modifications.
By contrast, \ourmethod achieves satisfactory structural distance (31.79$\downarrow$, second only to inversion-based methods), indicating superior strctural preservation during editing operations. Furthermore, our method maintains favorable background fidelity, performing comparably to leading inversion-free instruction-based editing methods (InstructPix2Pix~\cite{brooks2022instructpix2pix}, MGIE~\cite{fu2023guiding}) in PSNR and SSIM metrics. Notably, \ourmethod attains high CLIP similarity scores for both whole image and edited regions (24.19\&21.33, only second to PnPInversion and P2P), highlighting its ability to align edited content with target text prompts while preserving semantic consistency.

While certain diffusion-based methods outperform \ourmethod, these approaches are specifically optimized for diffusion frameworks. Our method pioneers training-free structural control within the AR paradigm, establishing a new technical pathway for AR image manipulation. 
% This breakthrough demonstrates that AR models can achieve comparable editing quality to diffusion counterparts without requiring architecture-specific training or inversion techniques.

\minisection{Generalizability.} Our method is generalizable across diverse AR base models. We include such qualitative experiments based on the Lumina-mgpt~\cite{liu2024lumina_mgpt} in the Supplementary.

\begin{figure}[t]
    \centering
\vspace{-3mm}
\includegraphics[width=0.999\columnwidth]{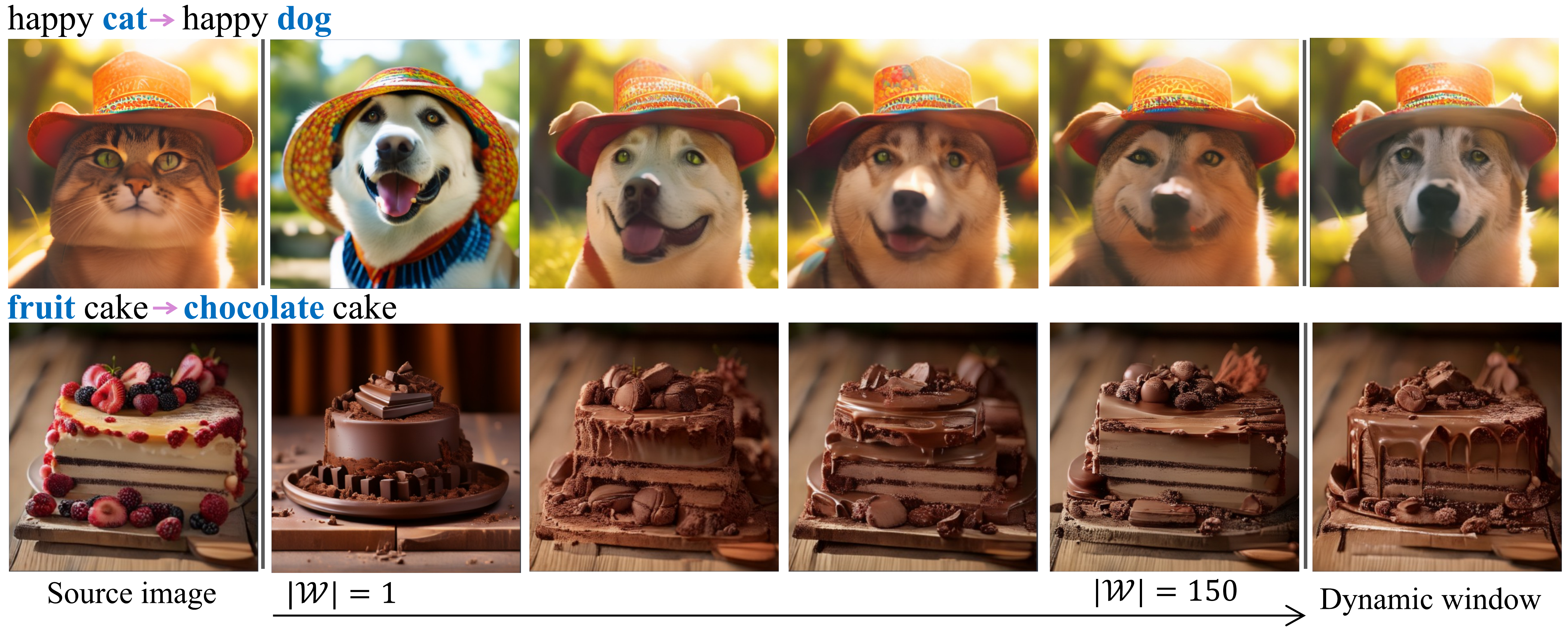}
\vspace{-7mm}
    \caption{The impact of window size $|\mathcal{W}|$. As the window size increases, structural preservation improves but flexibility decreases. Dynamic window strategies achieve better balance.    }
\vspace{-3mm}
    \label{fig:wsize_ablation}
\end{figure}

\begin{figure}[t]
    \centering    
    % \vspace{-3mm}
    \includegraphics[width=1\linewidth]{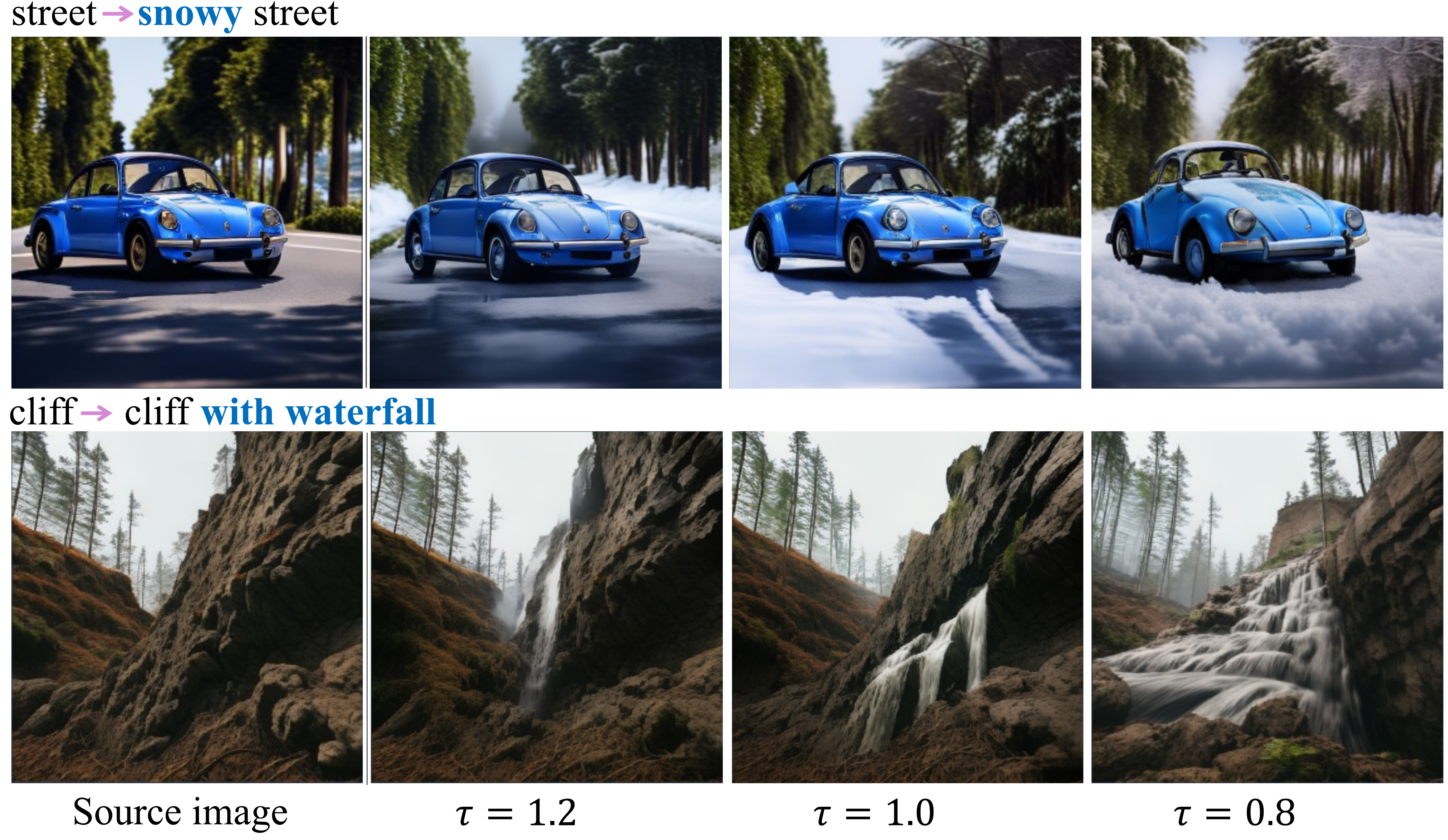}
\vspace{-6mm}
    \caption{Threshold $\tau$ affects the intensity of image editing. As it decreases, the editing effect becomes more pronounced, such as increased snow coverage and greater waterfall flow.}
\vspace{-3mm}
    \label{fig:tau_ablation}
\end{figure}

\begin{table}[t!]
\centering
\resizebox{0.9\columnwidth}{!}{
\begin{tabular}{cccc}
\hline
\textbf{Win. Size} & \textbf{Struc. Dist.$\downarrow$} & \textbf{Clip Sim.$\uparrow$} & \textbf{S/C}$\downarrow$ \\
\hline
$|\mathcal{W}|=50$ & 60.83 & \textbf{24.79} & 2.45 \\
$|\mathcal{W}|=100$ & 38.03 & 24.33 & 1.56 \\
$|\mathcal{W}|=150$ & \textbf{30.39} & 22.18 & 1.37 \\
Dynamic (Ours) &  \underline{31.79} & \underline{24.19} & \textbf{1.31} \\
\hline
\end{tabular}
}
% \caption{Effect of Window Size on Structure Distance and Clip Similarity}
\vspace{-3mm}
\caption{Ablation study over window size $|\mathcal{W}|$.}
\vspace{-6mm}
\label{tab:wsize_ablation}
\end{table}

\subsection{Ablation Study}
% To evaluate the importance of each component in \ourmethod, we conduct ablation studies by modifying one component at a time and observing its impact on the editing results.

\minisection{Effect of Window Size $|\mathcal{W}|$}.
% As shown in Fig.~\ref{fig:wsize_ablation} and Table~\ref{tab:wsize_ablation}, the ablation study on window size $|\mathcal{W}|$ reveals its critical role in balancing structural preservation and model generation flexibility. When $|\mathcal{W}| = 1$, our method degenerates into naïve prompt modification, resulting in less controlled edits. As $|\mathcal{W}|$ increases, the structural distance between the edited image and the original decreases, indicating better preservation of image composition. However, this comes at the cost of reduced generative flexibility. As seen in  Fig.~\ref{fig:wsize_ablation}, when $|\mathcal{W}|$ is too large, a \quotes{happy dog} with its tongue out is modified to have a closed mouth, and the chocolate coating on a cake is replaced with a plain bread layer. This effect is further reflected in the drop in CLIP Similarity (Table~\ref{tab:wsize_ablation}), suggesting weaker alignment with the target prompt.
% In contrast, our dynamic window strategy, shown in the last column of Fig.~\ref{fig:wsize_ablation}, achieves a better balance between structural consistency and semantic alignment, resulting in the best Structure Distance / CLIP Similarity ratio.
As shown in Fig.~\ref{fig:wsize_ablation} and Table~\ref{tab:wsize_ablation}, the ablation on window size $|\mathcal{W}|$ reveals a critical trade-off between structural preservation and generative flexibility. A larger $|\mathcal{W}|$ enhances structural preservation (lower structural distance) but can excessively constrain the model, thereby reducing generative flexibility and alignment with the target prompt (lower CLIP Similarity). For instance, an overly large window might incorrectly alter a dog's expression or a cake's topping, as seen in Fig.~\ref{fig:wsize_ablation}. In contrast, our proposed dynamic window strategy achieves a superior balance, resulting in the best Structure Distance / CLIP Similarity ratio.

\minisection{Effect of Threshold $\tau$}.
As illustrated in Fig.~\ref{fig:tau_ablation}, the threshold $\tau$ significantly influences the extent of modifications, particularly in high-variance editing tasks such as object addition. During image editing, there are cases that no candidate token closely matches the anchor token. 
Adjusting $\tau$ in these scenarios controls editing strength.
A lower $\tau$ allows more tokens from the original sampling to be retained, leading to more substantial modifications. This can be observed in Fig.~\ref{fig:tau_ablation}, where reducing $\tau$ results in thicker snow accumulation (1st row) and increased waterfall flow (2nd row). Conversely, a higher $\tau$ better preserves consistency with the original image, maintaining a more restrained edit.

\vspace{-1mm}
\section{Conclusion}
% \vspace{-1mm}
\label{sec:conclusion}
In this work, we address the fundamental challenge of text-guided image editing in autoregressive (AR) models without modifying model parameters or relying on explicit attention manipulation. 
By introducing Implicit Structure Locking (\ourmethod), we enable training-free image editing through a novel candidate matching protocol named Anchor Token Matching (\ouratm), which aligns the structure of the original image while preserving generative autonomy. 
% Unlike direct attention map transplantation, 
Our method \ourmethod ensures structure preservation as a natural consequence of latent space structure, allowing AR models to maintain spatial coherence for various editing tasks. 
Extensive experiments demonstrate that \ourmethod achieves competitive performance in structure-aware autoregressive editing, bridging the gap between AR models and diffusion models.
% bridging the gap between sequential token-based generation (AR models) and spatially controlled image synthesis (diffusion models). 
% This work represents the first step toward unlocking the full capability of AR models for high-fidelity, flexible image editing potentiality.

\section*{Acknowledgements}
% This work was supported by NSFC (NO. 62225604), Youth Foundation (62202243), and Shenzhen Science and Technology Program (JCYJ20240813114237048).
%
This work was supported by NSFC (NO. 62225604) and Youth Foundation (62202243).We acknowledge “Science and Technology Yongjiang 2035” key technology breakthrough plan project (2024Z120).  We also acknowledge the support of the Shenzhen Science
and Technology Program (JCYJ20240813114237048), the project PID2022-143257NB-I00 funded by the Spanish Government through MCIN/AEI/10.13039/501100011033 and FEDER, and the Generalitat de Catalunya CERCA Program. Computation is supported by the Supercomputing Center of Nankai University (NKSC).
%
% We acknowledge ``Science and Technology Yongjiang 2035'' key technology breakthrough plan project (2024Z120).
%
% Computation is supported by the Supercomputing Center of Nankai University (NKSC).

% \clearpage
% \clearpage

{
    \small
    \bibliographystyle{ieeenat_fullname}
    \bibliography{longstrings,main}
}
\clearpage
\appendix

\section{Implementation Details}

\subsection{Method Configuration}
Our implementation builds upon the official codebases of LlamaGen~\cite{sun2024autoregressive} and Lumina-mGPT~\cite{liu2024lumina_mgpt}. For LlamaGen-based editing, we employ a candidate window size of $K = 150$ and similarity threshold $\tau = 1.0$, utilizing Euclidean distance metrics in the VQ-AutoEncoder codebook's latent space. The Lumina-mGPT implementation adopts $K = 100$ and $\tau = 0.4$, measuring token distances through cosine similarity of first-layer transformer embeddings. All experiments were conducted on NVIDIA 3090 GPUs.

\subsection{Baseline Implementation}
For the diffusion-based methods we compared, including P2P~\cite{hertz2022prompt}, Null-text inversion~\cite{mokady2022null}, PnPInversion~\cite{direct_inversion_2023}, Pix2Pix-zero~\cite{parmar2023zero}, MasaCtrl~\cite{cao2023masactrl}, InstructPix2Pix~\cite{brooks2022instructpix2pix}, and MGIE~\cite{fu2023guiding}, we utilized their official implementations. For the two simple autoregressive (AR) model-based baselines we implemented:
\begin{itemize}
    \item \textbf{Naive Modify Prompt (\textit{NPM})}: This baseline modifies the original prompt to the edited prompt while keeping all other variables (e.g., non-edited words and random seeds) unchanged.
    \item \textbf{PnP-AR}: In this baseline, we save the token-wise and layer wise attention maps computed during the generation process of the original prompt. These attention maps are then directly replaced at the corresponding token positions and layers when generating images from the edited prompt.
\end{itemize}

\subsection{Evaluation Benchmarks}
Our method focuses on five fundamental editing types: object replacement, object addition, object deletion, style transfer, and attribute modification. For each editing type, we randomly select 10 examples from the corresponding category in the PIE-Bench~\cite{direct_inversion_2023} dataset. Each example includes an original prompt and an edited prompt. 

We first use LlamaGen to generate images based on the original prompts and then apply our method to edit these images according to the edited prompts. Due to the inherent limitations of LlamaGen in generating high-quality results from short prompts~\cite{sun2024autoregressive}, we employ GPT-4o mini~\cite{achiam2023chatgpt4} as a prompt enhancer to refine and improve the prompts before generation.

\section{Attention Map Analysis}
\subsection{Attention mechanism in AR}
\begin{figure}
    \centering
    \includegraphics[width=0.75\linewidth]{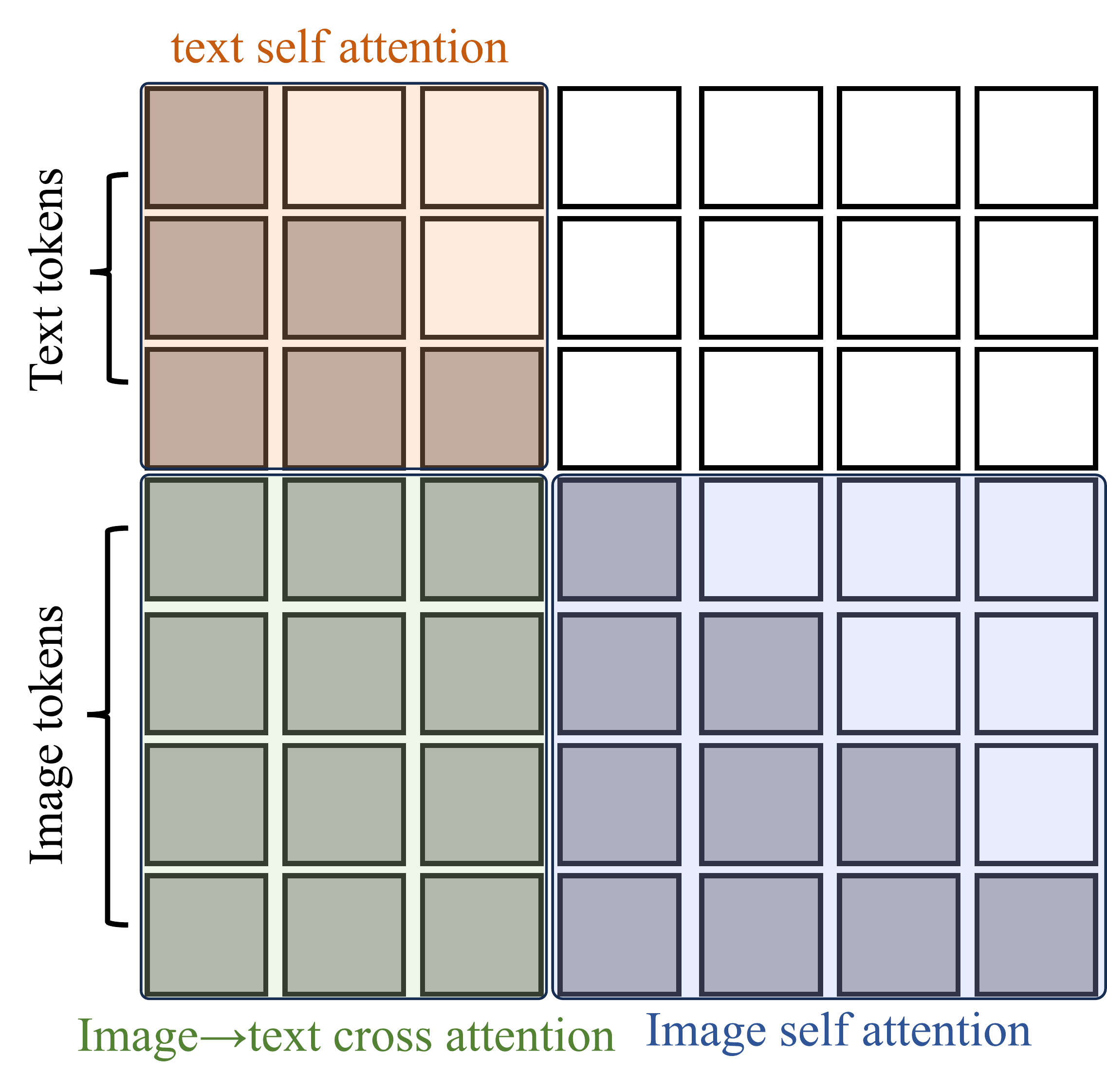}
    \caption{Illustration of the Attention mechanism in LlamaGen~\cite{Sun2024LLamaGen:AutoregressiveGeneration}.}
    \label{fig:attn_explain}
\end{figure}

As illustrated in Figure~\ref{fig:attn_explain}, for autoregressive image generation models such as LlamaGen, the text prompt is first encoded by the text encoder to obtain text tokens, which serve as the prefix tokens for the entire generation sequence. During the generation of each subsequent image token, attention is computed with both the preceding image tokens and the entire set of text tokens. The self-attention maps presented in our paper are derived from the image self-attention mechanism, as shown in the bottom-right section of Figure~\ref{fig:attn_explain}, while the cross-attention maps are obtained from the image-to-text cross-attention mechanism, depicted in the bottom-left section of Figure~\ref{fig:attn_explain}.

\subsection{Attention Visualization}
\begin{figure}[t]
    \centering
    \includegraphics[width=1\linewidth]{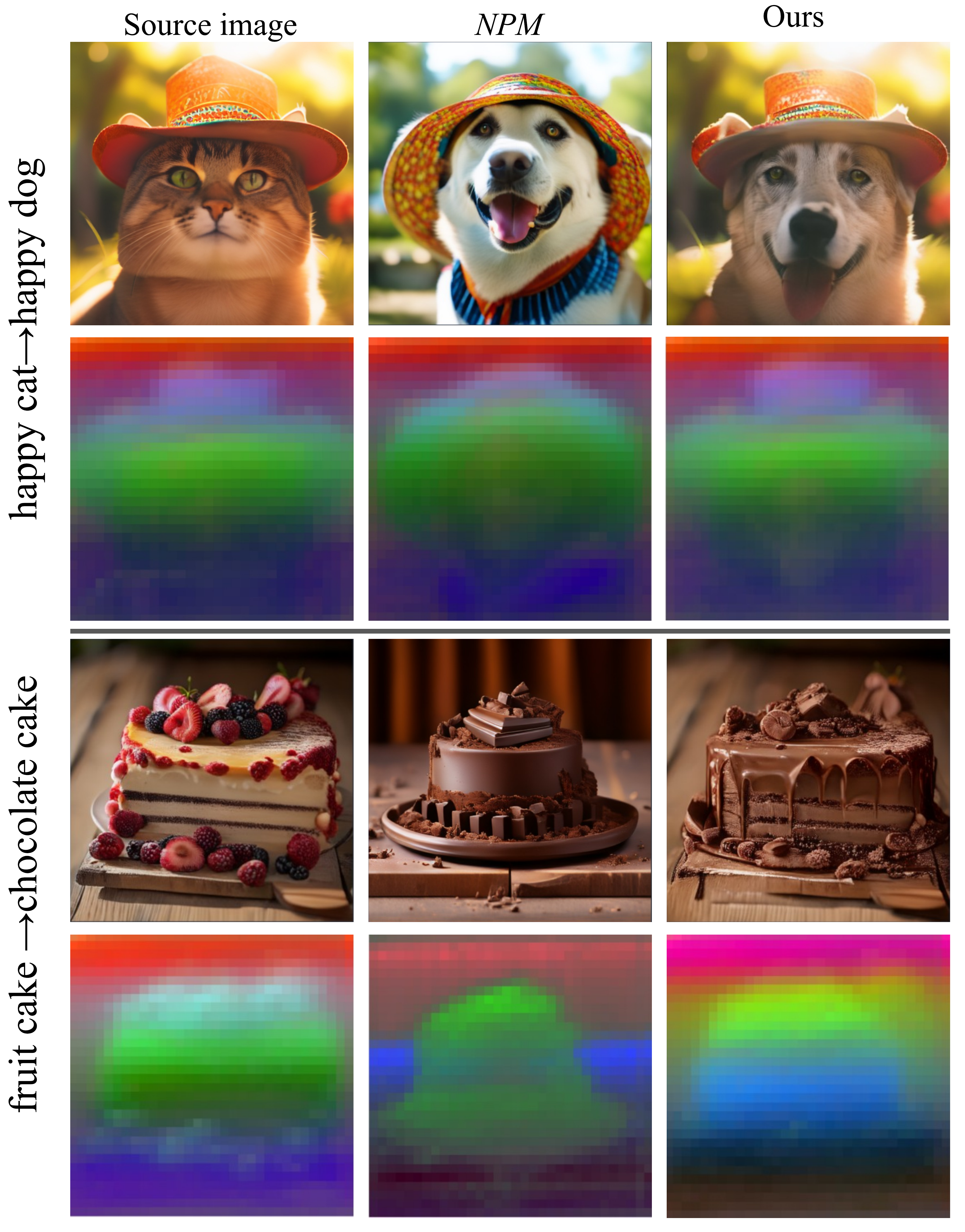}
    \caption{Ablation studies on attention maps. Compared with \textit{NPM}, our method \ourmethod naturally achieves better alignment of the original image and edited image generation processes during the attention map process.}
    \label{fig:attn_bala}
\end{figure}

Our method does not explicitly inject attention maps. Instead, structural preservation is implicitly achieved through anchor token matching, which naturally results in attention map consistency as a byproduct. As shown in Figure~\ref{fig:attn_bala}, compared to the attention maps obtained using the NPM method, the attention maps of the edited images generated by our method align naturally with those of the original images.

\subsection{Attention Locality}
\begin{figure}[t]
    \centering
    \includegraphics[width=1\linewidth]{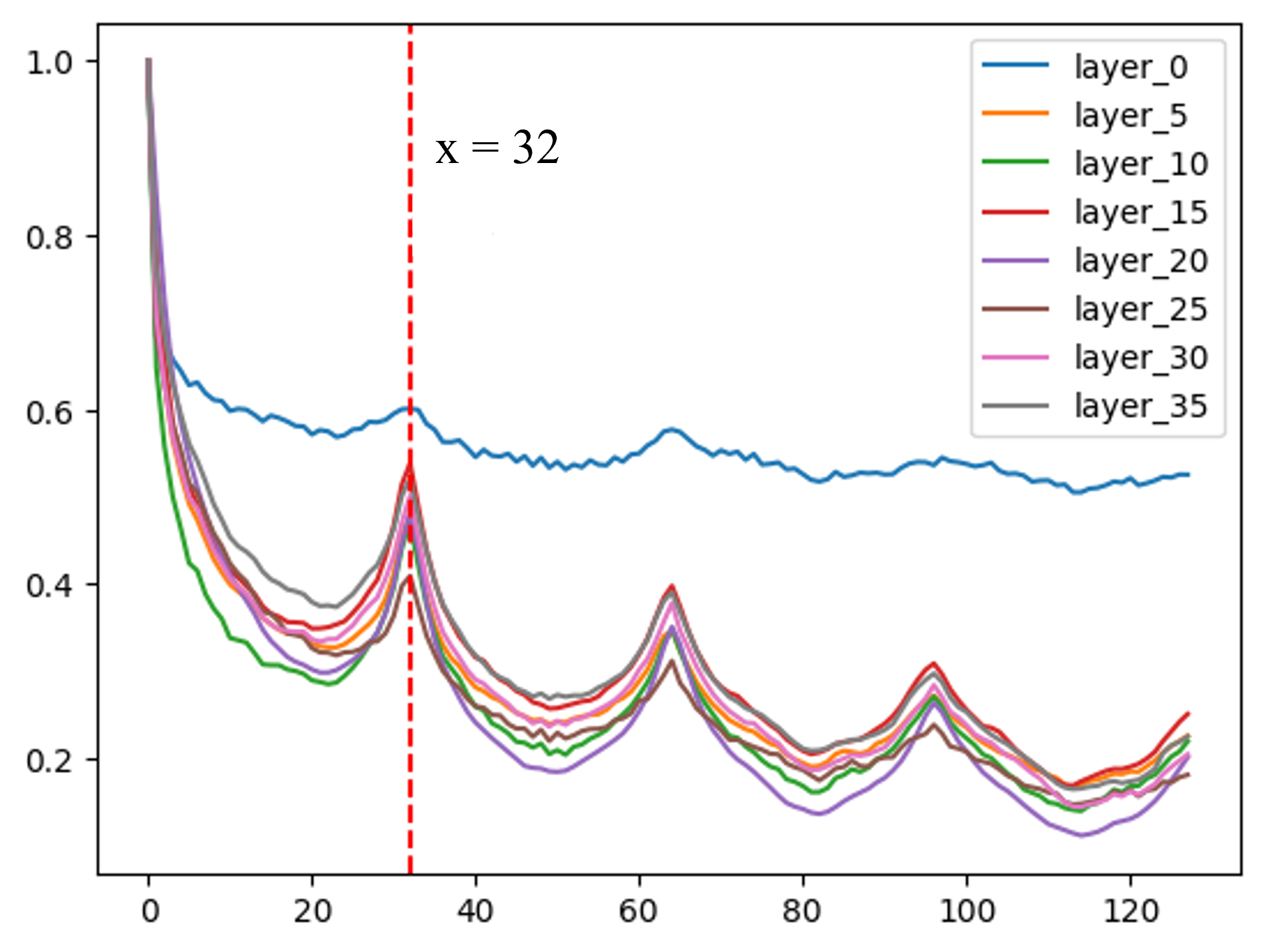}
    \caption{Autoregressive image generation models often tend to allocate larger attention score to tokens at adjacent positions.The values in the figure are normalized using min-max normalization.}
    \label{fig:attn_local}
\end{figure}
We observe that in autoregressive models such as LlamaGen, tokens tend to allocate higher attention weights to those adjacent to their positions during attention computation. As shown in Figure~\ref{fig:attn_local}, the attention score assigned by the current token decreases as the distance from the current token increases. However, the attention score periodically increases at intervals of 32 tokens. This phenomenon occurs because, when generating 512$\times$512 images, LlamaGen employs a VQ-Autoencoder to encode the image into a 32$\times$32 latent space. Tokens located at multiples of 32 positions away from the current token reside in the same column in the latent space, resulting in higher attention scores at these intervals.
This also explains why the cross-attention from image tokens to text tokens in Figure 4 of main paper shows that the earliest image tokens have the highest attention scores.

\section{More Visualization Comparisons}
\begin{figure*}
    \centering
    \includegraphics[width=1\linewidth]{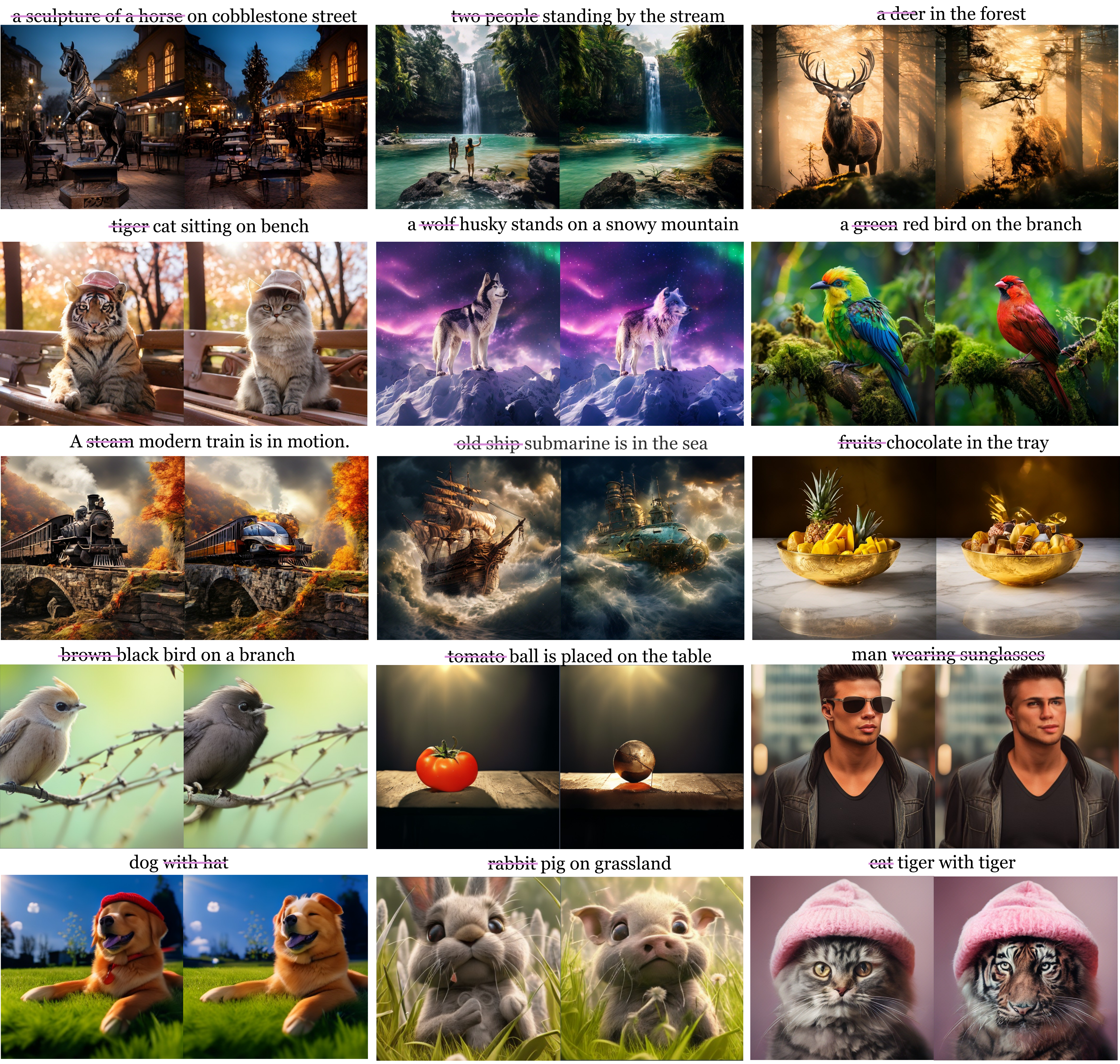}
    \caption{More visualization results of our method \ourmethod, where in the first three rows our method is integrated with lumina-mgpt~\cite{liu2024lumina_mgpt} and in the last two rows it is working with LlamaGen~\cite{Sun2024LLamaGen:AutoregressiveGeneration}.}
    \label{fig:more_rst}
\end{figure*}

In Figure~\ref{fig:more_rst}, we present additional editing results, demonstrating that our method generalizes well across different editing types and AR-based models.

\end{document}